\documentclass[letterpaper, 10 pt, conference]{ieeeconf}  

\IEEEoverridecommandlockouts                              

\usepackage{epsfig} 
\usepackage{times} 
\usepackage{amsmath} 
\usepackage{amssymb}  
\usepackage{graphicx}
\usepackage{hyperref}
\usepackage{xcolor}
\usepackage{multirow}
\usepackage{siunitx}
\usepackage{array}
\usepackage{diagbox}
\usepackage{adjustbox}
\usepackage{booktabs}
\usepackage{todonotes}
\usepackage{pifont}
\usepackage{amssymb}
\usepackage{url}
\usepackage{colortbl}
 
\newcommand{\greycell}{\cellcolor[RGB]{220,220,220}}%

\newcommand{\cmark}{\ding{51}}%
\newcommand{\xmark}{\ding{55}}%

\newcommand{\etal}{\emph{et~al.}}

\newcolumntype{P}[1]{>{\centering\arraybackslash}p{#1}}
\newcommand\crule[3][black]{\textcolor{#1}{\rule{#2}{#3}}}

\renewcommand{\baselinestretch}{0.985}

\newcolumntype{R}[2]{%
    >{\adjustbox{angle=#1,lap=\width-(#2)}\bgroup}%
    l%
    <{\egroup}%
}

\definecolor{sky}{RGB}{70, 130, 180}
\definecolor{road}{RGB}{70, 70, 70}
\definecolor{building}{RGB}{128, 64, 128}
\definecolor{sidewalk}{RGB}{244, 35, 232}
\definecolor{bicycle}{RGB}{190, 153, 153}
\definecolor{vegetation}{RGB}{107, 142, 35}
\definecolor{pole}{RGB}{255, 165, 0}
\definecolor{person}{RGB}{220, 20, 60}
\definecolor{car}{RGB}{0, 255, 0}
\definecolor{terrain}{RGB}{255, 255, 0}
\definecolor{fence}{RGB}{0, 255, 255}
\definecolor{curb}{RGB}{168, 168, 168}

\title{\LARGE \bf
HeatNet: Bridging the Day-Night Domain Gap in Semantic Segmentation with Thermal Images
}

\author{Johan Vertens$^{*}$, Jannik Z\"urn$^{*}$, and Wolfram Burgard
\thanks{$^{*}$These authors contributed equally. All authors are with the University of Freiburg, Germany. Wolfram
Burgard is also with the Toyota Research Institute, Los Altos, USA. Corresponding author: {\tt\small vertensj@informatik.uni-freiburg.de}}%
}

\begin{document}

\maketitle
\thispagestyle{empty}
\pagestyle{empty}

\begin{abstract}
    
    The majority of learning-based semantic segmentation methods are optimized for daytime scenarios and favorable lighting conditions.
    Real-world driving scenarios, however, entail adverse environmental conditions such as nighttime illumination or glare which remain a challenge for existing approaches.
    In this work, we propose a multimodal semantic segmentation model that can be applied during daytime and nighttime. To this end, besides RGB images, we leverage thermal images, making our network significantly more robust.
    We avoid the expensive annotation of nighttime images by leveraging an existing daytime RGB-dataset and propose a teacher-student training approach that transfers the dataset's knowledge to the nighttime domain. We further employ a domain adaptation method to align the learned feature spaces across the domains and propose a novel two-stage training scheme. Furthermore, due to a lack of thermal data for autonomous driving, we present a new dataset comprising over 20,000 time-synchronized and aligned RGB-thermal image pairs. In this context, we also present a novel target-less calibration method that allows for automatic robust extrinsic and intrinsic thermal camera calibration.
    Among others, we employ our new dataset to show state-of-the-art results for nighttime semantic segmentation.
    
\end{abstract}

\section{INTRODUCTION}
\label{sec:introduction}

Robust and accurate semantic segmentation of urban scenes is one of the enabling technologies for autonomous driving in complex and cluttered driving scenarios. Recent years have shown great progress in RGB image segmentation for autonomous driving ~\cite{zhao2017pyramid, cordts2016cityscapes}, which were predominantly demonstrated in favorable daytime illumination conditions. While the reported results demonstrate high accuracies on benchmark datasets \cite{cordts2016cityscapes, neuhold2017mapillary}, these models tend to generalize poorly to adverse weather conditions and low illumination levels present at nighttime. This constraint becomes especially apparent in rural areas where artificial lighting is weak or scarce. In autonomous driving, to ensure safety and situation awareness, robust perception in these conditions is a vital prerequisite.

Transfer learning and domain adaptation approaches aim at narrowing the domain gap between a source domain, where supervised learning from labelled data is possible, to a target domain, where labelled data is either sparse or not available. Such approaches, as demonstrated in~\cite{tzeng2014deep} or~\cite{zhang2017curriculum}, allow to adapt a given segmentation model to a different domain. These approaches, however, do not leverage a complementary modality such as thermal infrared images that can contain more relevant information to solve a given task in certain environmental conditions than a single modality would provide.

\begin{figure}
\centering
\includegraphics[width=\linewidth]{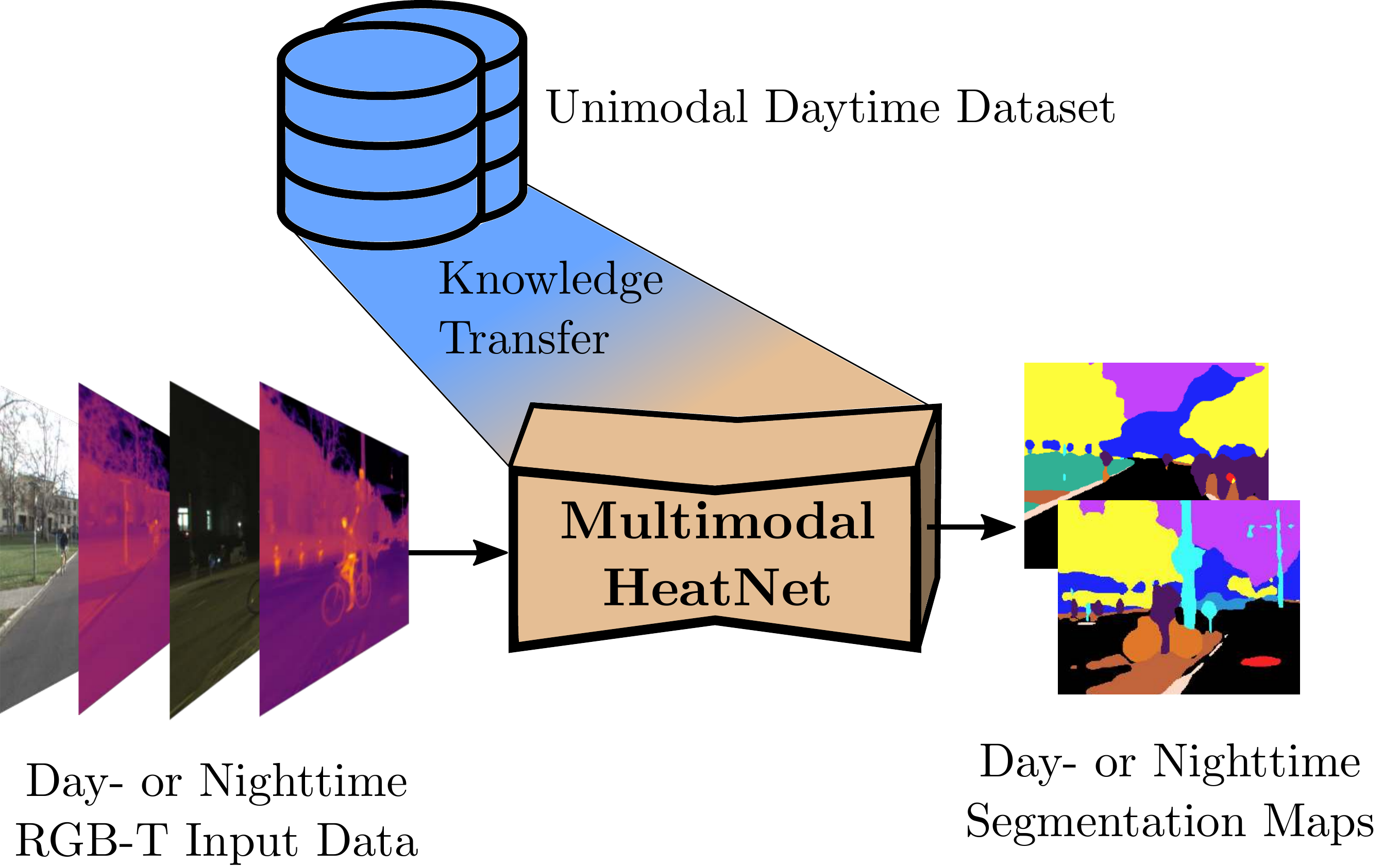}
\caption{Our multimodal segmentation network leverages both nighttime and daytime images. We transfer relevant knowledge from a large-scale unimodal daytime dataset for semantic segmentation with a teacher model to our multimodal HeatNet and simultaneously adapt our model to the nighttime domain by unsupervised domain adaptation.}
\label{fig:covergirl}
\end{figure}

In order to perform similarly well in challenging illumination conditions, it is beneficial for autonomous vehicles to leverage modalities complementary to RGB images \cite{valada2016deep, valada2017adapnet}. Encouraged by prior work in thermal image processing for object detection~\cite{wang2010improved}, object tracking~\cite{li2019rgb}, and semantic segmentation~\cite{ha2017mfnet, sun2019rtfnet}, we investigate leveraging thermal images for nighttime semantic segmentation of urban scenes. Thermal images contain accurate thermal radiation measurements with a high spatial density. Furthermore, thermal radiation is much less influenced by sunlight illumination changes and is less sensitive to adversary conditions. Existing RGB-thermal datasets for semantic image segmentation such as \cite{ha2017mfnet} are not as large-scale as their RGB-only counterparts. Thus, models trained on such datasets generalize poorly to challenging real-world scenarios.

In contrast to previous works, we utilize a semantic segmentation network for RGB daytime images as a teacher model to provide labels for the RGB daytime images in our dataset. We project the thermal images into the viewpoint of the RGB camera images using extrinsic and intrinsic camera parameters that we determine using our novel target-less camera calibration approach. Afterwards, we can reuse labels from this teacher model to train a multimodal semantic segmentation network on our daytime RGB-thermal image pairs. In order to encourage day-night invariant segmentation of scenes, we simultaneously train a feature discriminator
that aims at classifying features in the semantic segmentation network to belong either to daytime or nighttime images.

Furthermore, we propose a novel training schedule for our multimodal network that helps aligning the feature representations between day and night. Finally, we propose a new way of training a nighttime-daytime RGB-only semantic segmentation network by using thermal images as a bridge modality. In our baseline comparison and our ablation studies, we show that our model achieves comparable performance to fully supervised multimodal models. Additionally, we demonstrate that our first-of-its-kind method significantly reduces the domain gap between daytime and nighttime.


In summary, the contributions of this work are:

\begin{itemize}
	\item A novel multimodal approach for daytime and nighttime image segmentation, leveraging both RGB and thermal images while not requiring annotations for nighttime RGB or thermal infrared images.
    \item The \textit{Freiburg Thermal} dataset containing more than 20,000 time-synchronized RGB and thermal images recorded in urban and rural environments both in daytime and in nighttime conditions. We also provide \mbox{LiDAR} pointclouds, accurate GPS data and IMU readings.
    \item A novel target-less thermal camera calibration approach.   
    \item Extensive qualitative and quantitative evaluation of our approach, including ablation studies.
\end{itemize}

\section{RELATED WORKS}
\label{sec:relatedworks}

\subsection{Multimodal RGB-Thermal Datasets and Calibration}

While unimodal datasets with images in the visible domain are prevalent in computer vision research, some datasets have been proposed that entail aligned RGB-thermal image pairs. 
Berg \etal~\cite{berg2015thermal} propose a dataset that consists of thermal infrared images which is mainly targeted towards object tracking. Similarly, Li \etal~\cite{li2019rgb}, propose a RGB-thermal dataset for multimodal object tracking in varying outdoor settings and conditions.
The authors of CATS \cite{treible2017cats} present a general outdoor dataset for color and thermal stereo disparity estimation. Besides RGB-thermal image pairs, LiDAR-based ground-truth disparity maps are available.
Furthermore, the work of Shivakumar \etal~\cite{shivakumar2019pst900} targets the scenarios of the DARPA Subterranean Challenge providing 894 RGB-thermal image pairs with pixel-wise semantic annotations for underground rescue scenarios.
There exist only a few datasets that contain thermal infrared imagery in the context of autonomous driving. In the work of Hwang \etal~\cite{hwang2015multispectral}, a dataset is proposed that consists of more than 95k RGB-thermal image pairs. Each pair is annotated with bounding boxes for persons and is hence aimed towards pedestrian detection research.
The KAIST multispectral dataset \cite{choi2018kaist} entails multiple modalities such as RGB, thermal infrared, LiDAR, GNSS and IMU for a total of 7512 frames. They also provide annotations/ground-truth for 2D bounding boxes, drivable region, image enhancement, depth, and colorization.
The authors of MFNet \cite{ha2017mfnet} present the first urban scene dataset for multimodal semantic segmentation, comprising 1569 pixel-wise annotated RGB-thermal image pairs. Approximately half of the recorded images were captured during nighttime. However, many of the most common classes in the context of semantic segmentation for autonomous driving such as road, sidewalk, pole, sign, building or sky are not annotated.

Due to the lack of large-scale RGB-thermal datasets for urban semantic segmentation, we propose the Freiburg Thermal dataset comprising over 20000 high-resolution RGB-thermal image pairs in particularly challenging environments. We additionally provide semantic annotations for a distinct test set.

For most previously proposed datasets, distinct RGB and thermal cameras were used and calibrated leveraging hand-made patterns such as checkerboards \cite{treible2017cats, shivakumar2019pst900} or lines on printed circuit boards \cite{choi2018kaist}.
A different approach was presented by Lussier \etal~\cite{lussier2014automatic} in which an edge response map between depth and thermal images is minimized using grid search over the calibration parameter space. 

In contrast to prior work, we propose a method to calibrate the intrinsic, extrinsic and distortion parameters of the thermal infrared camera in a purely target-less fashion, leveraging spatial transformer networks~\cite{jaderberg2015spatial} and stochastic gradient descent over a large number of images.

\subsection{Semantic Segmentation of Thermal Images}

Recently, semantic segmentation of thermal images began to attract more attention in the computer vision community. Qiao~\etal~\cite{qiao2017thermal} use a level set method to detect pedestrians in thermal images. More recently, Li~\etal~\cite{li2019segmenting} proposed an edge-conditioned segmentation network for thermal images, trained supervised on a dataset containing various indoor and outdoor scenes. The works closest to our work are \cite{ha2017mfnet} and \cite{sun2019rtfnet}. In the work of Ha \etal~\cite{ha2017mfnet}, the authors propose a multimodal fusion network architecture for RGB and thermal images. They evaluate their approach on their own dataset MF ~\cite{ha2017mfnet}. Similarly, Sun~\etal~\cite{sun2019rtfnet} propose an RGB-thermal fusion network and show their results on the MF dataset.

In contrast to the works mentioned above, we train an RGB-thermal semantic segmentation model without requiring any manual labeling efforts. We instead use a teacher model trained solely on RGB images to provide supervision for the daytime image pairs.
We further present an extended multimodal domain adaptation method that enables robust nighttime segmentation.

\subsection{Domain Adaptation for Semantic Segmentation}

Many works in transfer learning explore unsupervised domain adaptation from synthetic data to real environments \cite{tsai2018learning}, \cite{chen2018road}, \cite{yang2019adversarial}. Other recent works explore model adaptation from daytime to nighttime via an intermediate twilight domain \cite{dai2018dark}, \cite{sakaridis2019guided}. Following a different approach, works were proposed that conduct unpaired image-to-image translation using generative models to create synthetic nighttime training data \cite{porav2019don, sun2019see, romera2019bridging}. Most similar to our work, in \cite{wulfmeier2017addressing}, the authors investigate adversarial domain adaptation, where they use a binary classifier to discriminate between daytime and nighttime image features produced by an encoder network. A domain confusion loss penalizes features that can easily be classified as originating from the daytime or nighttime domain.

In contrast to the above works, our approach leverages additional modalities such as thermal images that provide complementary inputs for semantic segmentation in challenging illumination conditions, significantly narrowing the daytime-nighttime domain gap. 


\section{TECHNICAL APPROACH}
\label{sec:approach}


In the following, we describe our approach to multimodal semantic segmentation for daytime and nighttime scenes, leveraging RGB and thermal images. 
In our approach, we first train a semantic segmentation teacher model in a supervised fashion on the Mapillary Vistas dataset~\cite{neuhold2017mapillary}. Subsequently, we use this teacher network to infer labels of daytime RGB images on our multimodal Freiburg Thermal dataset. We then train a student network supervised on the daytime image annotations provided by the teacher model, using both RGB and thermal infrared images. While the thermal modality is mostly invariant to lighting changes, the RGB modality differs significantly between daytime and nighttime and thus exhibits a significant domain gap. We thus further utilize a domain adaptation technique that aligns the internal feature distributions of the multimodal segmentation network, enabling the network to perform similarly well for nighttime images as for daytime images. As thermal cameras are not yet available in most autonomous platforms, we further propose to distill the knowledge from the domain-adapted multimodal model back into a unimodal segmentation network that exclusively uses RGB images.

In the following we detail our approach.

\subsection{RGB-T Semantic Segmentation}
\label{sec:rgbtsemseg}

\begin{figure*}
\centering
\includegraphics[width=\linewidth]{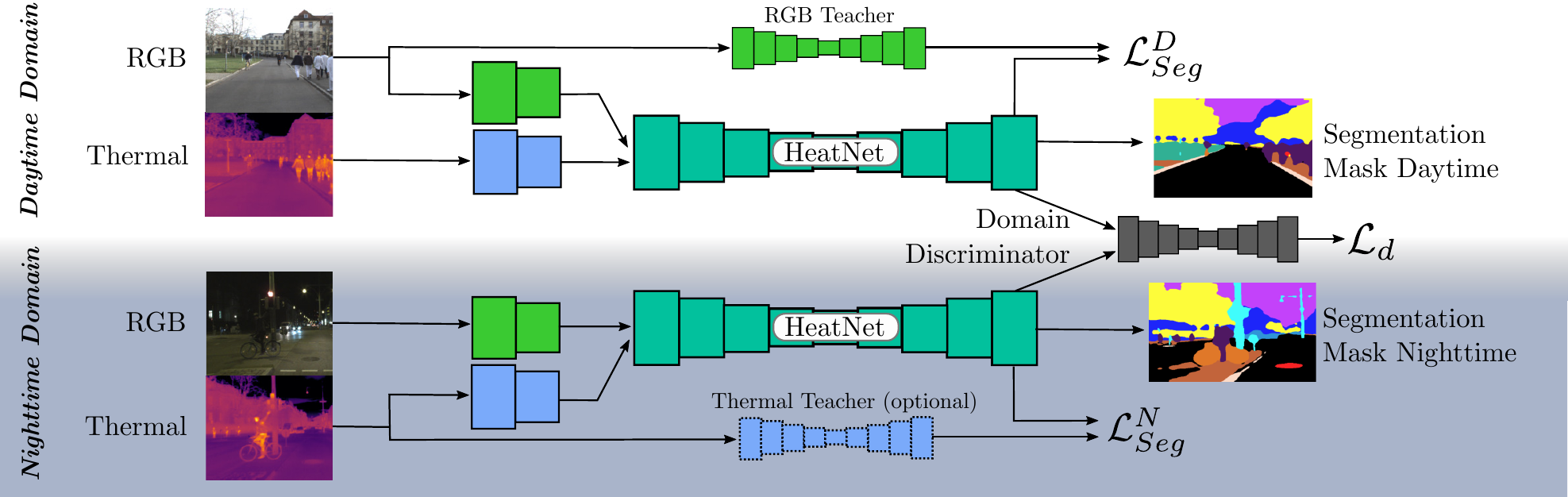}
\caption{Our proposed HeatNet architecture uses both RGB and thermal images and is trained to predict segmentation masks in daytime and nighttime domains. We train our model with daytime supervision from a pre-trained RGB teacher model and with optional nighttime supervision from a pre-trained thermal teacher model trained on exclusively thermal images. We simultaneously minimize the cross entropy prediction loss to the teacher model prediction and minimize a domain confusion loss from a domain discriminator to reduce the domain gap between daytime and nighttime images.}
\label{fig:architecture}
\end{figure*}

We initially train a PSPNet model~\cite{zhao2017pyramid} for semantic RGB image segmentation on the Mapillary Vistas dataset~\cite{neuhold2017mapillary}, which contains 20,000 RGB images and semantic annotations from highly diverse and challenging urban scenes. We use this model as a teacher model $M_D$ for daytime images, providing pixel-wise semantic annotations for all daytime RGB images in our Freiburg Thermal dataset. Since we project each thermal image into the viewpoint of the RGB camera using the extrinsic and intrinsic camera calibration parameters, described in Sec. \ref{sec:camera_calib}, we can use the same annotations for each respective thermal image. Given the labels produced by $M_D$, we subsequently train our multimodal RGB-T model $M_M$ by minimizing the cross-entropy loss between the network and the teacher model prediction. Note that the teacher model can only provide supervision for the daytime domain since we can assume that $M_D$ does not generalize to nighttime images as it is not trained on data from this domain. We formulate the daytime segmentation loss as:
\begin{equation}
\begin{aligned}
    \mathcal{L}_{s}^{D}  = - \frac{1}{H W} \sum_{h,w} M_D(I_{\mathrm{RGB}}^{D}) \log M_M(I_{\mathrm{RGB}}^{D}, I_{\mathrm{T}}^{D}),
\end{aligned}
\label{eq:lsd}
\end{equation}

where $M_D(I_{\mathrm{RGB}}^{D})$ denotes the teacher model prediction and $M_M(I_{\mathrm{RGB}}^{D}, I_{\mathrm{T}}^{D})$ denotes the prediction of our multimodal RGB-T model for daytime thermal images $I_\mathrm{T}^D$ and RGB images $I_\mathrm{RGB}^D$. $H$ and $W$ denote the height and width of the output, respectively. For simplicity, we omit the class index $i$ in Eq. \ref{eq:lsd}. By supervised training using the labels from $M_D$, the student model $M_M$ does not generalize well to nighttime scenes because of the large domain shift in the RGB domain.
In order to adapt the model to the nighttime domain in an unsupervised manner, we utilize a domain adaptation approach similar to \cite{tsai2018learning} and insert a domain discriminator $C$ after the softmax prediction layer of $M_M$. The domain discriminator has as inputs the softmax activations $S_D$ or $S_N$ of our segmentation model for daytime or nighttime inputs, respectively, and is trained to differentiate between both domains. We thus define the discriminator loss $\mathcal{L}_{d}$ as

\begin{equation}
    \mathcal{L}_{d} = \frac{1}{H W} \sum_{h,w}
    \begin{cases}
    [0 - C(S_X)]^2,& \text{if } X = D\\
    [1 - C(S_X)]^2,& \text{if } X = N 
\end{cases}
\end{equation}

In order to adapt our model to the nighttime domain we aim to predict semantic segmentation maps that fool the discriminator model. In other words, we want to output predictions whose origin is classified as the daytime domain. If this confusion of the discriminator model can be achieved, it can be assumed that the distribution of the internal feature representations of our multimodal model are matched and the model is adapted to the nighttime domain. We train our model with an alternating training scheme for the two networks, where we step-wise alternate between adjusting the parameters of the discriminator model while freezing the segmentation model parameters and adjusting the parameters of the segmentation model while freezing the discriminator model parameters. In each iteration we sample an RGB-T image pair from the daytime and nighttime domain. In the first step of an iteration, we train our semantic segmentation network for the daytime domain while adapting the nighttime feature representations to daytime. We minimize an overall loss $\mathcal{L}_{p_1}$:

\begin{equation}
\begin{aligned}
    \mathcal{L}_{p_1}  = \mathcal{L}_{s}^{D} + \lambda [0 - C(S_N)]^2,
\end{aligned}
\label{eq:Lp1}
\end{equation}

where $\lambda$ denotes a constant weighting factor between both losses, which we set to $0.01$ during all experiments. In the second step, we exclusively train the discriminator to differentiate between day and night segmentation maps with the overall loss $\mathcal{L}_{p_2}$: 

\begin{equation}
\begin{aligned}
    \mathcal{L}_{p_2}  = \mathcal{L}_{d}
\end{aligned}
\end{equation}

Our model architectures and the overall training scheme is illustrated in Fig. \ref{fig:architecture}. In addition to the described approach, we propose the following extensions:

\subsubsection{Two-Stage Training}
\label{sec:thermal_night_init}

We argue that the domain gap between day and night is much smaller for thermal images than for RGB images. This results in superior nighttime performance if a network is exclusively trained on thermal infrared images without any domain adaptation. 
As our goal is to train a multimodal network that performs best in both domains, day and night, we conduct domain adaptation to compensate for the illumination changes in the RGB images. We argue, however, that during domain adaptation the training could converge into local minima due to the large domain gap within the RGB modality and insufficient feature distribution overlap.
We thus propose to first train our multimodal network $M_M$ with the daytime teacher model $M_D$ and an additional nighttime teacher model $M_N$. This additional teacher network is trained only on thermal infrared images and therefore predicts reasonable nighttime segmentation maps without domain adaptation. Furthermore, we argue that the semantic annotations provided by $M_N$ can still be improved. Therefore, after the training with both teacher networks, we continue with the normal training procedure including domain adaptation, but without the nighttime teacher model, which we explained in the previous section. Following this training scheme, the feature representations align reasonably in the first training stage and the domain adaption in the second training stage does not need to bridge the full domain gap anymore.

\subsubsection{RGB-T to RGB Model Distillation}
\label{sec:rgb_distillation}

Since thermal infrared cameras are not always installed on mobile robots, we propose a simple, yet effective strategy to enable RGB-only nighttime semantic segmentation using our approach. As previously mentioned, due to the domain gap in the visible spectrum, it is challenging to adapt an RGB-only model to the nighttime domain. Meanwhile, with our previous multimodal adaption approach we are capable of training a multimodal network that leverages RGB information jointly with thermal infrared information which exhibits a significant smaller domain gap.
We thus propose to first train a multimodal RGB-T network following the previously described method. We afterwards distill the knowledge of the RGB-T network to an RGB-only network. To this end we use the previously described RGB-only daytime teacher model to provide supervision in daytime and our best-performing RGB-T network to provide supervision in nighttime and train this RGB-only model fully supervised in both domains.

\section{DATASET}
\label{sec:dataset}

\begin{figure}
\centering
\includegraphics[width=\linewidth]{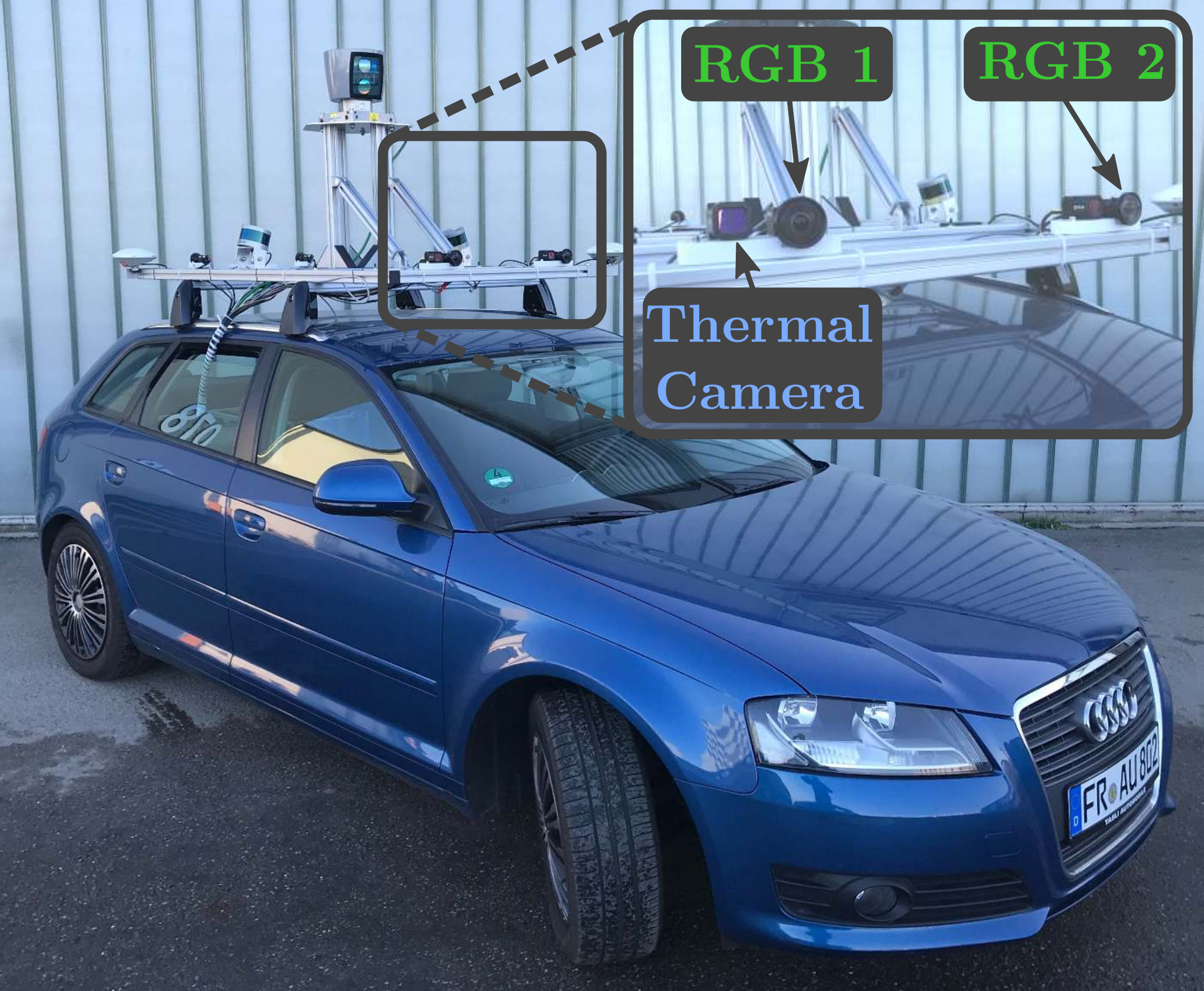}
\caption{Our stereo RGB and thermal camera rig mounted on our data collection vehicle.}
\label{fig:cameras}
\end{figure}

To kindle research in the area of thermal image segmentation and to allow for credible quantitative evaluation, we create the large-scale dataset \textit{Freiburg Thermal}. We provide the dataset and the code publicly available at \href{http://thermal.cs.uni-freiburg.de/}{\textit{http://thermal.cs.uni-freiburg.de/}}. The Freiburg Thermal dataset was collected during 5 daytime and 3 nighttime data collection runs, spanning the seasons summer through winter. Overall, the dataset contains 12051 daytime and 8596 nighttime time-synchronized images using a stereo RGB camera rig (FLIR Blackfly 23S3C) and a stereo thermal camera rig (FLIR ADK) mounted on the roof of our data collection vehicle. In addition to images, we recorded the GPS/IMU data and LiDAR point clouds. The Freiburg Thermal dataset contains highly diverse driving scenarios including highways, densely populated urban areas, residential areas, and rural districts.
We also provide a testing set comprising 32 daytime and 32 nighttime annotated images. Each image has pixel-wise semantic labels for 13 different object classes. Annotations are provided for the following classes: \textit{Road, Sidewalk, Building, Curb, Fence, Pole/Signs, Vegetation, Terrain, Sky, Person/Rider, Car/Truck/Bus/Train, Bicycle/Motorcycle,} and \textit{Background}. We deliberately selected extremely challenging urban and rural scenes with many traffic participants and changing illumination conditions.

\subsection{Camera calibration}
\label{sec:camera_calib}
For our segmentation approach it is important to perfectly align RGB and thermal images as otherwise the RGB teacher model predictions would not be valid as labels for the thermal modality. Thus, in order to accurately carry out the camera calibration for the thermal camera, we propose a novel target-less calibration procedure. While in previous works \cite{shivakumar2019pst900,luhmann2013geometric} different  kinds of checkerboards or circleboards have been leveraged, our method does not require any pattern. Although, for RGB cameras, these patterns can be produced and utilized easily, it still remains a challenge to create patterns that are robustly visible both in RGB and thermal images. In general, the used modalities infrared and RGB entail different information. However, we note that the edges of most common objects in urban scenes are easily observable in both modalities. Thus, in our approach we minimize the pixel-wise distance between such edges. In the case of aligning two monocular cameras, targetless calibration without any prior information results in ambiguities for the estimation of the intrinsic camera parameters. We therefore utilize our pre-calibrated RGB stereo rig in order to provide the missing sense of scale. Due to the target-less nature of our approach, our thermal camera calibration method can be easily deployed in an online calibration scenario.

%
%

Our aim is to overlay the RGB and thermal images as best as possible, solving both for the extrinsic and intrinsic parameters. If this alignment can be achieved, our cameras are assumed to be fully calibrated. In the following we assume the RGB image $I_{\mathrm{RGB}}$ to be undistorted and rectified. We formulate the misalignment $E$ as the difference between the gradients of the calibrated RGB image and the transformed thermal image as:

\begin{equation}
\begin{aligned}
    E = \sum_{u,v} [\nabla I_{\mathrm{RGB}} - \nabla \mathcal{S}(I_{T}, F)]
\end{aligned}
\end{equation}

Here, $\mathcal{S}(I_{\mathrm{T}}, F)$ denotes a function that transforms a source thermal image $I_{\mathrm{T}}$ to a target RGB image $I_{\mathrm{RGB}}$ while using a pixel displacement map $F$ that maps from $I_{\mathrm{T}}$ to $I_{\mathrm{RGB}}$. A successful calibration would result in the minimum value of $E$ and would therefore align the thermal image with the RGB image. We follow \cite{jaderberg2015spatial} in order to implement $\mathcal{S}$, using differentiable spatial transformer networks.

We compute $F$ by projecting the pixel coordinates of the RGB images to 3D, transforming them into the thermal camera coordinate system and projecting them back to the thermal image plane. Thus, the displacement map $F=p_{\mathrm{RGB}} -p_{\mathrm{T}}$ between the RGB pixel coordinates $p_{\mathrm{RGB}}$ and the thermal image pixel coordinates $p_{\mathrm{T}}$ can be found with:

\begin{equation}
\begin{aligned}
    p_{RGB} = \phi\big(K_{\mathrm{T}} \: T_{{\mathrm{RGB} \rightarrow \mathrm{T}}} \gamma (p_\mathrm{RGB} \mid K_{\mathrm{RGB}},D_{\mathrm{RGB}})\big)
\end{aligned}
\end{equation}

where the function $\gamma (p \mid K, D) = D(p) K^{-1} h(p)$ backprojects the RGB pixel coordinate into the 3D camera coordinate system while $h(p)$ transforms $p$ in the homogeneous vector form. The intrinsic calibration of the RGB camera is denoted as $K_{\mathrm{RGB}}$ and $D_{\mathrm{RGB}}$ refers to the depth corresponding to the RGB image $I_{\mathrm{RGB}}$. Further, $T_{{\mathrm{RGB} \rightarrow \mathrm{T}}}$ and $K_{\mathrm{T}}$ refer to the sought extrinsic and intrinsic thermal camera calibration values, respectively. The function $\phi(x)$ simply divides the vector $x$ by its last element. We infer $D_{\mathrm{RGB}}$ by leveraging a dense stereo depth estimation method based on a convolutional neural network \cite{chang2018pyramid}. Due to the locality of the edges within the RGB and the thermal image, the direct minimization of the misalignment $E$ would lead to vanishing gradients and would prevent fast convergence on the global minimum. In order to cope with this problem, we convolve the difference of gradients $\big(\nabla I_{\mathrm{RGB}} - \nabla \mathcal{S}(I_{\mathrm{T}}, F)\big)$ with a large Gaussian kernel $G(\sigma)$ with zero mean, standard deviation $\sigma=3$, and $51$ pixel aperture size, resulting in our loss function:

\begin{equation}
\begin{aligned}
\label{eq.cal_loss}
    \mathcal{L}_{c} = \sum_{u,v} \big[G(\sigma) * \big( \nabla I_{\mathrm{RGB}} - \nabla U(\mathcal{S}(I_{\mathrm{T}}, F), v)\big)\big]^2
\end{aligned}
\end{equation}

We follow \cite{heikkila1997four} to model the distortion of the thermal image by the function $U$ and optimize its parameters  $v = [k_1$, $k_2$, $p_1$, $p_2]$, referring to radial and tangential distortion respectively, while optimizing the objective function. 

We define the extrinsic calibration $T_{{\mathrm{RGB} \rightarrow \mathrm{T}}}$ as a rigid body transformation $T_{{\mathrm{RGB} \rightarrow \mathrm{T}}} = \begin{pmatrix}R & t\\ 0 & 1\end{pmatrix} \in \mathit{SE}(3)$ where $R \in \mathit{SO}(3)$ and $t \in \mathbb{R}^3$. In order to ease the optimization we optimize the transformation in Lie-algebraic exponential coordinates $\xi = (v^T \: \omega^T) \in \mathfrak{se}(3)$ and use the exponential map with small-angle approximations \cite{eade2013lie} to map from $\mathfrak{se}(3)$ to $\mathit{SE}(3)$.

In our implementation we use Adam \cite{kingma2014adam} for stochastic gradient descent to minimize Eq. \ref{eq.cal_loss} which yields the optimal extrinsic calibration $T_{{\mathrm{RGB} \rightarrow \mathrm{T}}}^*$, thermal camera intrinsic matrix $K_{\mathrm{T}}^*$, and undistortion parameters $v^*$.

We take 600 random image-pairs for the optimization process and set a batch size of 10. Furthermore, we set the number of iterations to 8000 and halve the step size every 500 steps. We initialize $K_{\mathrm{T}}$ as:
\begin{equation}
\begin{aligned}
\label{eq.cam_matrix}
    K_{\mathrm{T}}=\begin{pmatrix} f_m/l & 0 & r_w / 2\\ 0 & f_m/l & r_h/2 \\ 0 & 0 & 1 \end{pmatrix},
\end{aligned}
\end{equation}
where $f_m$ denotes the ideal manufactured focal-length of the lens, $l$ the size of a single square pixel in mm, $r_w$ the horizontal resolution and $r_h$ the vertical resolution. All other parameters such as extrinsic calibration and distortion parameters are set to $10^{-4}$ to prevent vanishing gradients.

\begin{figure}
\centering
\includegraphics[width=\linewidth]{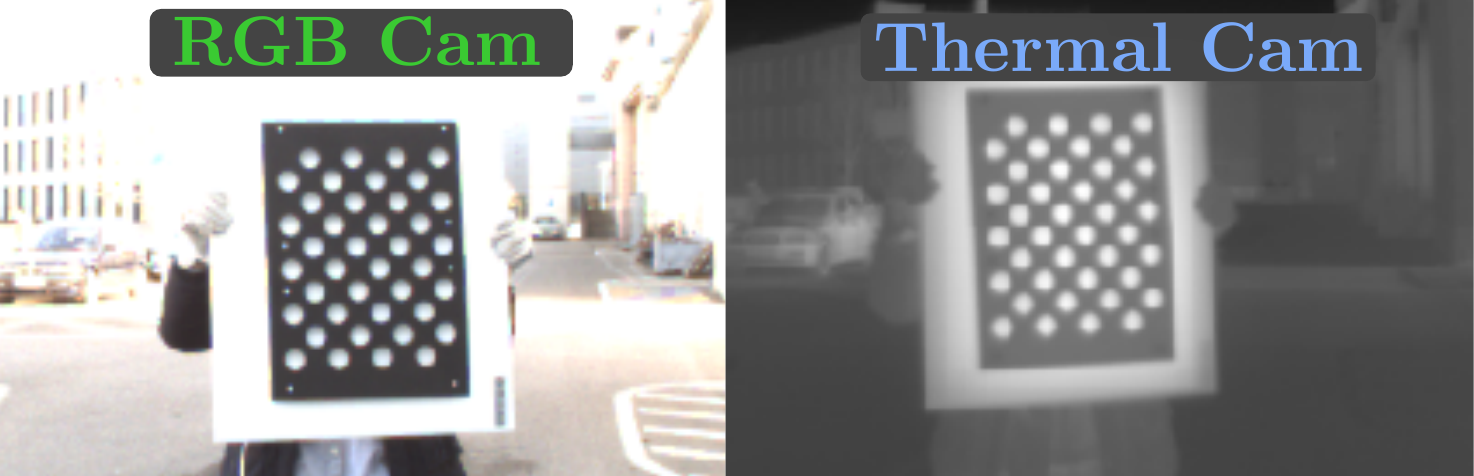}
\caption{Our calibration board placed in front of a heating panel is equally well visible in the RGB and thermal domain. The recorded image-pairs are used to obtain the calibration parameters with Kalibr~\cite{furgale2013unified}.}
\label{fig:cal_board}
\end{figure}

We qualitatively compare the RGB-thermal image alignment obtained with our target-less approach to a circleboard-based calibration procedure carried out using the publicly available tool Kalibr \cite{furgale2013unified}. We manufactured a circleboard and placed it in front of a heating panel. Fig. \ref{fig:cal_board} shows our calibration board as recorded by the RGB and thermal camera. The recorded image-pairs were used to obtain the extrinsic and intrinsic calibration parameters with Kalibr.


\begin{figure}
\scriptsize 
\centering 
\setlength{\tabcolsep}{0.3em}
\renewcommand{\arraystretch}{1}
\includegraphics[width=.99\linewidth]{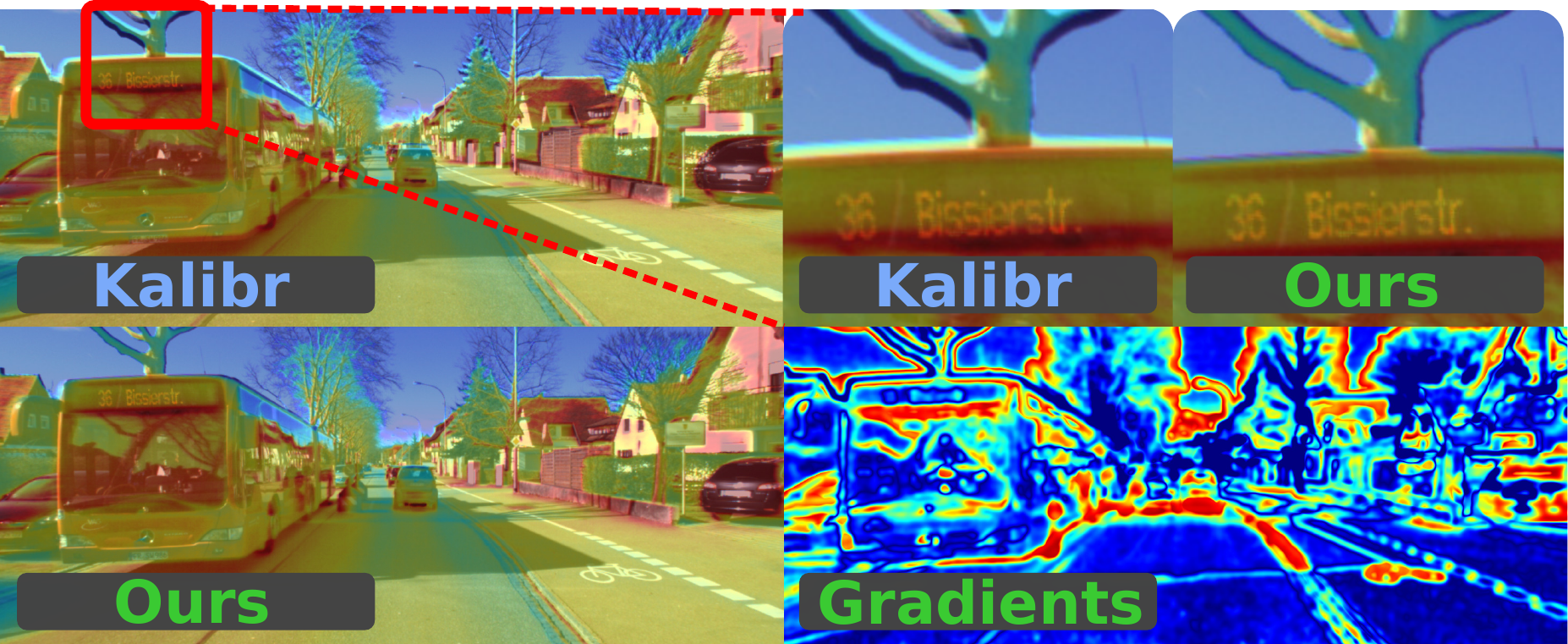}
\caption{Qualitative result of our target-less RGB-T calibration approach. In the left column, we show the RGB and thermal image alignment overlay with calibration parameters as obtained with Kalibr and our approach, respectively. The magnified view in the top-right corner demonstrates that our approach yields superior alignment of object edges. The bottom-right corner illustrates the magnitude of gradient difference between RGB and thermal image after the optimization process with our approach.}
\label{fig:cal_examples}
\vspace{-0.3cm}
\end{figure}

Fig. \ref{fig:cal_examples} qualitatively compares the RGB-thermal alignment obtained with our target-less approach to the alignment obtained with calibration parameters produced by Kalibr. Despite our approach not requiring any calibration targets, we observe that our approach yields qualitatively better alignment of RGB and thermal images.

\section{Experimental Results}
\label{sec:experiments}

In the following we present the experimental results of our proposed multimodal semantic segmentation method. We evaluate our model on our proposed \textit{Freiburg Thermal} dataset and on MF \cite{ha2017mfnet}.
Furthermore, we present results on the 30 nighttime images of the Berkeley Deep Drive dataset \cite{yu2018bdd100k}, using the unimodal RGB version of HeatNet, leveraging our proposed knowledge distillation approach described in Sec.~\ref{sec:rgb_distillation}. We also present various ablation studies and provide a discussion of all results.

\subsection{Network Architecture}
As our unimodal architecture for the teacher networks $M_D$ and $M_N$ we use the PSPNet architecture \cite{zhao2017pyramid}. For our multimodal network we again adopt the PSPNet architecture but replicate the first two blocks of the corresponding ResNet-50 encoder. After passing the individual modalities through the replicated blocks we concatenate the feature maps and proceed with the remaining blocks of the encoder. For the discriminator architecture we follow the described architecture in \cite{tsai2018learning}.

\subsection{Training Details}

We train our HeatNet segmentation model for $100$ epochs with the RMSprop optimizer and with an initial learning rate of $10^{-4}$. We use learning rate halving every $30$ epochs. In each training batch, using our alternating training scheme, we forward the RGB-T image pair and minimize Eq. \ref{eq:Lp1}. We set the batch size to $8$ for all our experiments.

\subsection{Baseline Comparison}

We report the performance of HeatNet trained on Freiburg Thermal and tested on Freiburg Thermal, MF, and on the BDD night test split. All results are listed in Tab.~\ref{tab:baselineSegComp}. We observe that our RGB Teacher model $M_D$, which is trained on the Vistas dataset \cite{neuhold2017mapillary}, has a high mIoU score of $69.4$ in the day domain and an expected low score of $35.7$, as the network is neither trained nor adapted to the night domain. Our thermal teacher model $M_N$ achieves a mIoU score of $57.0$, which shows that the domain gap is much smaller for this domain as for RGB. Our final RGB-T HeatNet model achieves with $64.9$ the overall best score on our test set. Furthermore the RGB-only HeatNet reaches a comparable score to our RGB-T variant, proving the efficiency of our distillation approach which leverages the thermal images as a bridge modality.

We deploy the same distilled RGB network to publish results on the night BDD split. It can be observed that our method boosts the mIoU by $50\%$.


In order to compare the performance of our network with the recent RGB-T semantic segmentation approaches MFNet \cite{ha2017mfnet} and RTFNet-50~\cite{sun2019rtfnet}, we also fine-tune our model on the $784$-image MF \cite{ha2017mfnet} training set and report scores on the corresponding test set. We select all classes that are compatible between MF and Freiburg Thermal for evaluation which are the classes \textit{Car}, \textit{Person}, and \textit{Bike}. 
We train our method only with labels provided by the teacher model $M_D$, while not requiring any nighttime labels or labels from MF in general. Thus, it is expected that MFNet and RTFNet outperform HeatNet as they are trained supervisedly. However, it can be observed that HeatNet achieves comparable numbers to MFNet.


We further evaluate the generalization properties of the models trained on MF and tested on our FR-T dataset. We observe that the model performance deteriorates when evaluating MFNet or RTFNet on our FR-T dataset. We conclude that the diversity and complexity of the MF dataset does not suffice to train robust and accurate models for daytime or nighttime semantic segmentation of urban scenes.

\begin{figure*}
\centering
\footnotesize
\setlength{\tabcolsep}{0.1cm}
    \begin{tabular}{P{0.7cm}P{2.6cm}P{2.6cm}P{2.6cm}P{2.6cm}P{2.6cm}P{2.6cm}}
        Dataset & RGB Image & Thermal Image & RGB-Teacher & HeatNet RGB & HeatNet RGB-T & Ground Truth \\
           FR-T & \raisebox{-.5\height}{\includegraphics[width=\linewidth]{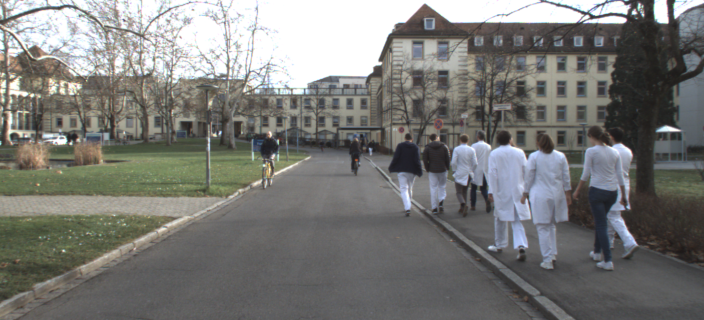}} & \raisebox{-.5\height}{\includegraphics[width=\linewidth]{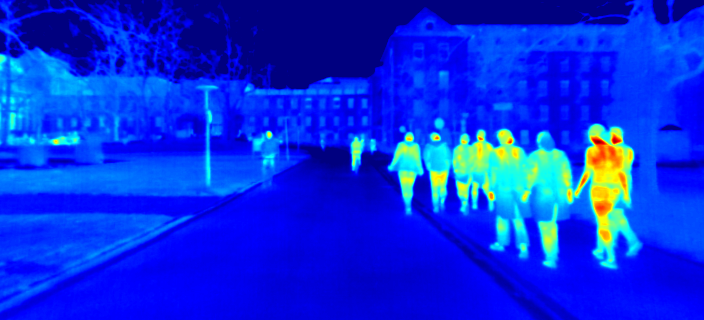}} &  \raisebox{-.5\height}{\includegraphics[width=\linewidth]{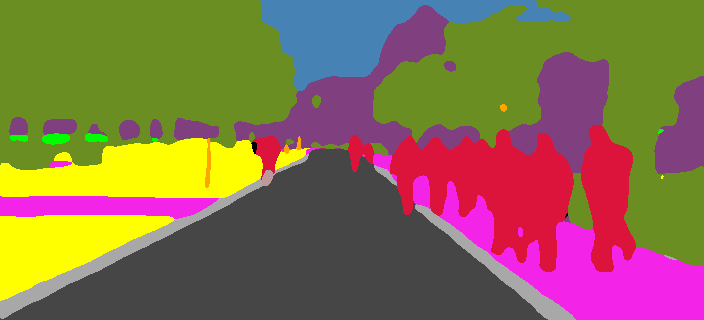}} & \raisebox{-.5\height}{\includegraphics[width=\linewidth]{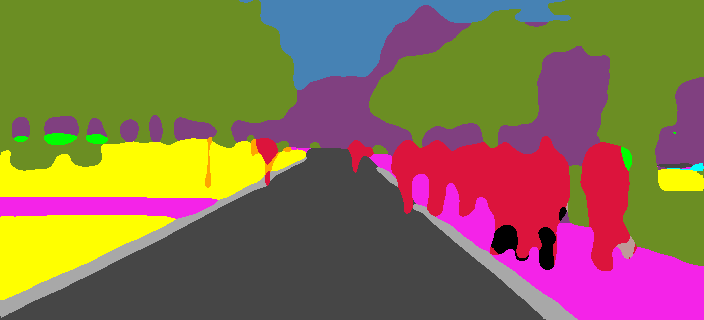}} & \raisebox{-.5\height}{\includegraphics[width=\linewidth]{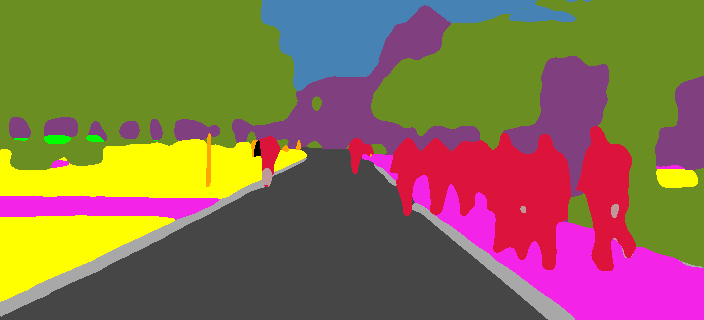}} & \raisebox{-.5\height}{\includegraphics[width=\linewidth]{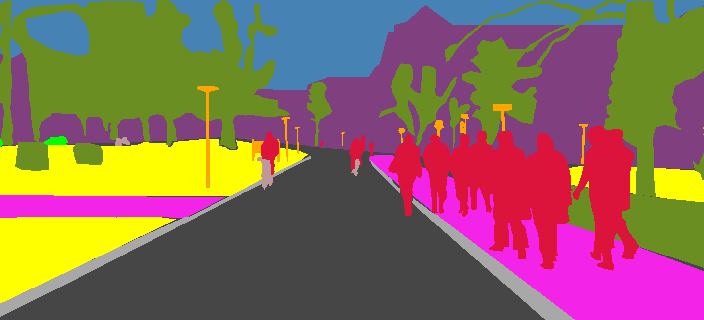}}\\
          FR-T & \raisebox{-.5\height}{\includegraphics[width=\linewidth]{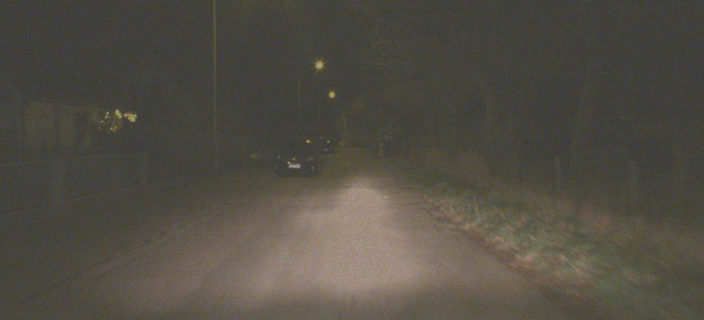}} &  \raisebox{-.5\height}{\includegraphics[width=\linewidth]{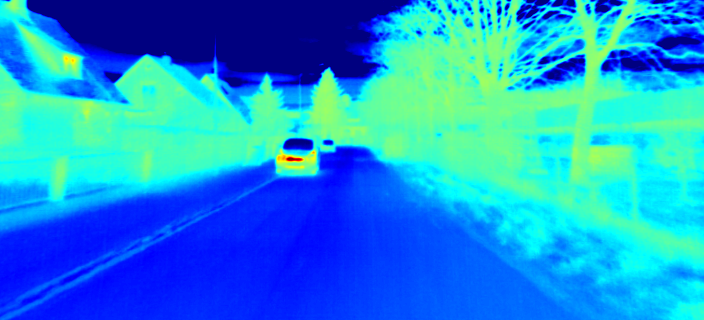}} & \raisebox{-.5\height}{\includegraphics[width=\linewidth]{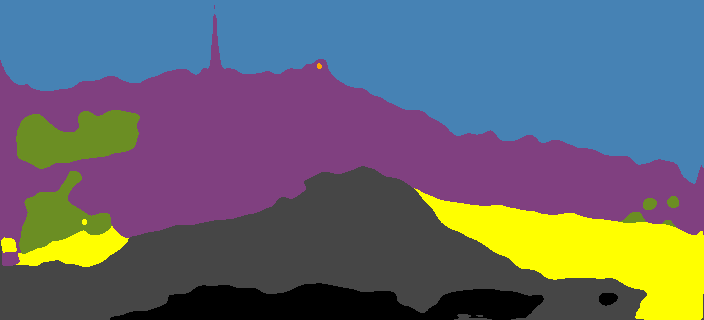}} & \raisebox{-.5\height}{\includegraphics[width=\linewidth]{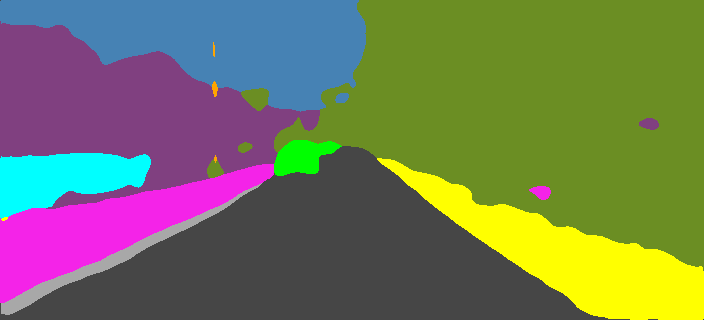}} & \raisebox{-.5\height}{\includegraphics[width=\linewidth]{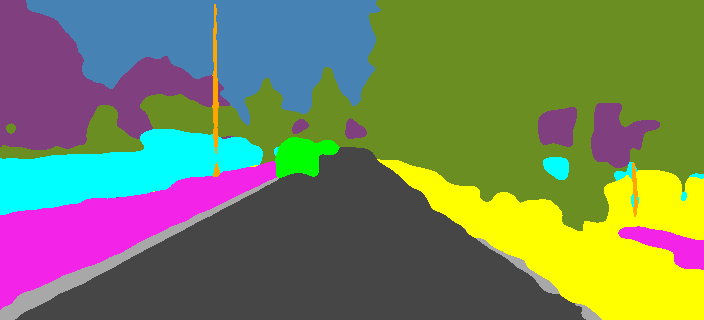}} & \raisebox{-.5\height}{\includegraphics[width=\linewidth]{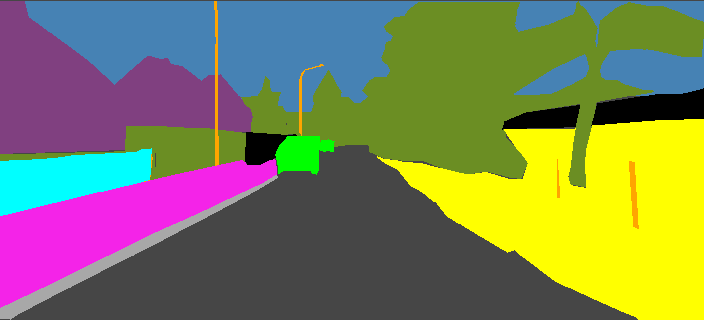}}\\
          \noalign{\smallskip}\hline{\smallskip}
          & RGB Image & N/A  & RGB Teacher & HeatNet RGB & N/A & Ground Truth\\
          BDD & \raisebox{-.5\height}{\includegraphics[width=\linewidth]{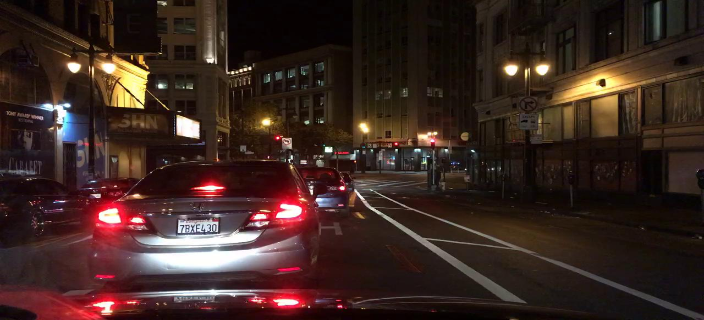}} & \raisebox{-.5\height}{\includegraphics[width=\linewidth]{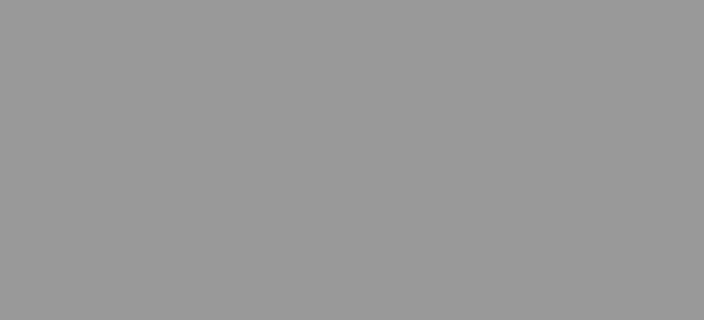}} & \raisebox{-.5\height}{\includegraphics[width=\linewidth]{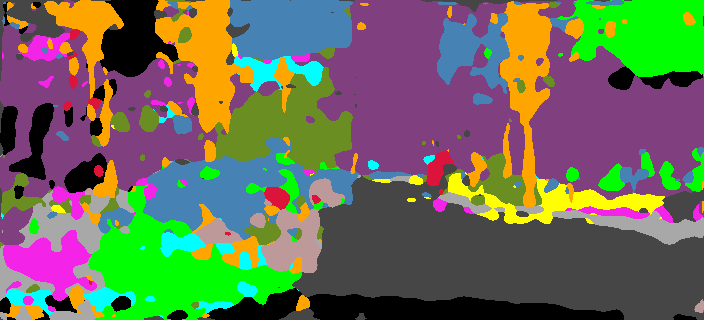}} & \raisebox{-.5\height}{\includegraphics[width=\linewidth]{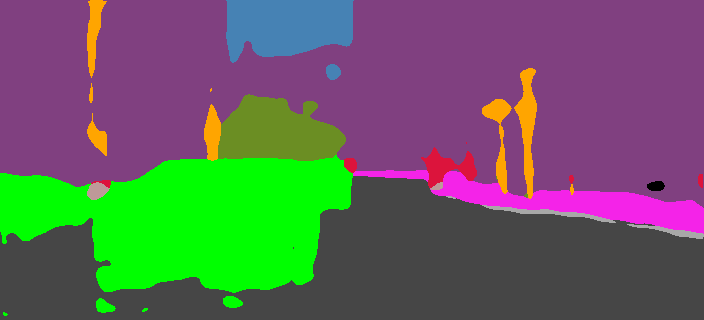}} & \raisebox{-.5\height}{\includegraphics[width=\linewidth]{figures/no_image.png}} & \raisebox{-.5\height}{\includegraphics[width=\linewidth]{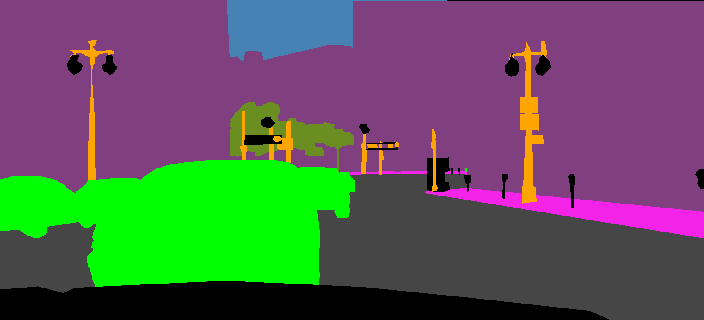}}\\
    \end{tabular}
    \caption{Qualitative semantic segmentation results of our model variants. We compare segmentation masks of our RGB-only teacher model, HeatNet RGB-only, and HeatNet RGB-T to ground truth. In the first two rows, we show segmentation masks obtained on the Freiburg Thermal dataset. The bottom row illustrates results obtained on the RGB-only BDD dataset. The multimodal approaches cannot be evaluated on BDD and the corresponding images are left blank.}
    \label{fig:qualitative} 
\end{figure*}

\begin{table*}
\scriptsize  
\centering
\caption{Comparison of RGB-Thermal semantic segmentation performance with state-of-the-art approaches on the MF dataset and on the Freiburg Thermal (FR-T) dataset. We mark results obtained using fully supervised methods with a gray background. Classes available for evaluation due to incompatible or missing annotations are marked with a dash (-).}
\label{tab:baselineSegComp}
\renewcommand{\arraystretch}{0.2}  
\begin{tabular}{p{1.0cm}|p{1.7cm}|p{2.0cm}p{0.25cm}p{0.25cm}p{0.33cm}p{0.33cm}p{0.33cm}p{0.33cm}p{0.33cm}p{0.33cm}p{0.33cm}p{0.33cm}p{0.33cm}p{0.33cm}p{0.33cm}p{0.33cm}|p{0.33cm}}
\hline\noalign{\smallskip}
 &   &  &  &  & \rotatebox[origin=c]{90}{Road} & \rotatebox[origin=c]{90}{Sidewalk} & \rotatebox[origin=c]{90}{Building} & \rotatebox[origin=c]{90}{Curb} & \rotatebox[origin=c]{90}{Fence} & \rotatebox[origin=c]{90}{Pole} & \rotatebox[origin=c]{90}{Vegetation} & \rotatebox[origin=c]{90}{Terrain} & \rotatebox[origin=c]{90}{Sky} & \rotatebox[origin=c]{90}{Person} & \rotatebox[origin=c]{90}{Car} & \rotatebox[origin=c]{90}{Bicycle} &   \\
\centering{Train On} & \centering{Test On} & \centering{Model} & \centering{RGB} & \centering{T} & \crule[road]{0.2cm}{0.2cm} & \crule[sidewalk]{0.2cm}{0.2cm} & \crule[building]{0.2cm}{0.2cm} & \crule[curb]{0.2cm}{0.2cm} & \crule[fence]{0.2cm}{0.2cm} & \crule[pole]{0.2cm}{0.2cm} & \crule[vegetation]{0.2cm}{0.2cm} & \crule[terrain]{0.2cm}{0.2cm} & \crule[sky]{0.2cm}{0.2cm} & \crule[person]{0.2cm}{0.2cm} & \crule[car]{0.2cm}{0.2cm}  & \crule[bicycle]{0.2cm}{0.2cm}  & Mean  \\

\noalign{\smallskip}\hline\hline\noalign{\smallskip}

\centering{MF}   & \centering{MF}  & MFNet \cite{ha2017mfnet} &\cmark &\cmark & - & - & - & - & - & -  & - & - & - & \greycell 58.9 & \greycell 65.9 & \greycell 42.9 &  \greycell 55.9  \\
 &  &  RTFNet-50 \cite{sun2019rtfnet} & \cmark & \cmark & - & - & - & - & - & -  & - & - & - & \greycell \textbf{67.8} & \greycell \textbf{86.3} & \greycell \textbf{58.2} &  \greycell \textbf{70.7}  \\
 &  & HeatNet  & \cmark & \cmark & - & - & - & - & - & -  & - & - & - & 56.4 & 68.8 & 33.9 & 53.0  \\

\cmidrule{2-18}

& \centering{FR-T Day/Night}  & MFNet \cite{ha2017mfnet} & \cmark & \cmark & - & - & - & - & - & -  & - & - & - & \greycell 42.8 & \greycell 27.0 & \greycell 24.5 & \greycell 31.4  \\
&      & RTFNet-50 \cite{sun2019rtfnet} & \cmark & \cmark & - & - & - & - & - & -  & - & - & - & \greycell \textbf{63.2} & \greycell 61.5 & \greycell 51.3 & \greycell 58.6  \\
&      & HeatNet                        &\cmark  & \cmark & 86.7 & 57.5 & 67.7 & 46.4 & 41.5 & 43.8  & 57.9 & 44.1 & 63.7 & 63.1 & \textbf{85.6} & \textbf{58.2} & 59.7                 \\

\noalign{\smallskip}\hline\hline\noalign{\smallskip}

\centering{FR-T} & \centering{MF} & HeatNet & \cmark & \cmark & - & - & - & - & - & - & - & - & - & 51.6 & 61.8 & 30.2 & 47.9  \\

\cmidrule{1-18}

\centering{(Vistas)} & \centering{FR-T Day}   & RGB Teacher      & \cmark & \xmark & \textbf{89.7} &  \textbf{67.0} & 73.8 & 56.9 & 48.8 & 53.8 & 73.8 & 62.8 & 84.3 & 72.0 & 90.1 & 60.4 & 69.4  \\
\centering{FR-T} & & HeatNet & \cmark &  \cmark  & 89.4 & 65.6 & \textbf{74.8} & \textbf{59.7} & \textbf{52.9} & \textbf{54.3} & \textbf{74.1} & \textbf{65.1} & \textbf{84.5} & \textbf{74.0} & \textbf{91.2} & \textbf{64.1} & \textbf{70.8} \\  
\cmidrule{1-18}
\centering{FR-T} & \centering{FR-T Night} & Thermal Teacher  & \xmark & \cmark & 84.9 & 60.5 &  \textbf{65.5} & 43.1 & 31.8 & 38.1 & 51.8 & 40.1 & 72.6 & 49.6 & \textbf{87.1} & \textbf{56.9} & 57.0  \\
\centering{(Vistas)} &  & RGB Teacher      & \cmark & \xmark & 76.3 & 22.6 & 53.4 & 10.8 & 14.1 & 31.6 & 10.4 & 13.5 & 47.7 & 28.0 & 74.3 & 45.2 & 35.7  \\
\centering{FR-T} &  & HeatNet  & \cmark &  \cmark  & \textbf{86.4} & \textbf{60.9} & 65.4 & \textbf{45.5} & \textbf{35.5} & \textbf{42.0} & \textbf{52.5} & \textbf{52.3} & \textbf{73.9} & \textbf{54.9} & 85.7 & 53.3 & \textbf{59.0} \\ 
\cmidrule{1-18}
\centering{FR-T} & \centering{FR-T Day/Night}  & HeatNet         & \cmark &  \cmark & \textbf{87.9} & \textbf{63.3} & \textbf{70.1} & \textbf{52.6} & \textbf{44.2} & \textbf{48.2} & \textbf{63.3} & \textbf{58.9} & \textbf{79.2} & \textbf{64.5} & \textbf{88.5} & \textbf{58.7} & \textbf{64.9} \\
\centering{FR-T} &   & HeatNet  RGB-only       & \cmark &  \xmark & 82.7 &  56.0 & 66.0 & 45.3 & 34.0 & 37.8 & 58.4 & 49.5 & 71.0 &  54.4 & 84.2 & 57.4 & 58.0  \\
\cmidrule{1-18}

\centering{(Vistas)} & \centering{BDD Night~\cite{yu2018bdd100k}}  & RGB Teacher        & \cmark  & \xmark   & 68.8 &  21.5 & 32.9 & - & 0.0 & 12.3 & 11.5 & 6.6 & \textbf{27.2} &  24.5 & 40.4 & - & 24.6  \\
\centering{FR-T} &   & HeatNet RGB-only  & \cmark & \xmark & \textbf{87.1} &  \textbf{40.0} & \textbf{50.2} & - & \textbf{25.9} & \textbf{22.9} & \textbf{12.8} & \textbf{8.5} & 25.0 & \textbf{27.4} & \textbf{68.3} & - & \textbf{36.8}  \\
\noalign{\smallskip}\hline\noalign{\smallskip}
\end{tabular}
\vspace{-0.4cm}
\end{table*}

\subsection{Ablation Studies}

In order to evaluate the various components of our HeatNet approach, we perform ablation studies with different variants of our model. All ablation studies presented in this section were performed on our Freiburg Thermal dataset and are listed in Tab.~\ref{tab:ablationStudy}.

\begin{table}
\footnotesize 
\centering
\caption{Ablation studies for variants of our HeatNet model on the Freiburg Thermal dataset.}
\label{tab:ablationStudy}
\renewcommand{\arraystretch}{0.3}  
\begin{tabular}{m{0.8cm} | p{0.3cm} p{0.3cm} p{1.0cm} p{1.0cm} | p{0.4cm} | p{0.5cm} | p{0.4cm} p{0cm} } 
\hline\noalign{\smallskip}
  & & & & & \multicolumn{3}{c}{\textbf{mIoU}}  \\
\centering{Variant} & \centering{RGB} & \centering{T} & \centering{Domain Discriminator} & \centering{Two-Stage Training}  & \centering{Day} & \centering{Night} & \centering{Both}  & \\

\noalign{\smallskip}\hline\hline\noalign{\smallskip}

\centering{V1} & \centering{\xmark}  & \centering{\cmark} & \centering{\xmark} &  \centering{\xmark}  & \centering{68.1} & \centering{57.0} & \centering{62.6}  &\\ 
\centering{V2} & \centering{\cmark}  & \centering{\xmark} & \centering{\xmark} &  \centering{\xmark}  & \centering{68.3} & \centering{25.1} & \centering{46.7}  &\\ 
\centering{V3} & \centering{\cmark}  & \centering{\cmark} & \centering{\xmark} &  \centering{\xmark}  & \centering{67.9} & \centering{33.7} & \centering{50.8}  &\\ 
\centering{V4} & \centering{\cmark}  & \centering{\xmark} & \centering{\cmark} &  \centering{\xmark}  & \centering{70.5} & \centering{43.2} & \centering{56.9}  &\\ %
\centering{V5} & \centering{\cmark}  & \centering{\cmark} & \centering{\cmark} &  \centering{\xmark}  & \centering{70.6} & \centering{56.3} & \centering{63.5}  &\\ 
\centering{V6} & \centering{\cmark}  & \centering{\cmark} & \centering{\cmark} &  \centering{\cmark}  & \centering{\textbf{70.8}} & \centering{\textbf{59.0}} & \centering{\textbf{64.9}}  &\\

\noalign{\smallskip}\hline\noalign{\smallskip}
\end{tabular}
\vspace{-0.6cm}
\end{table}


We first study the impact of the image modalities on the model performance without using domain adaptation or two-stage training. We compare a unimodal RGB-only model (V1) with a unimodal thermal-only model (V2) and the multimodal variant trained both on RGB and on thermal images (V3). All variants are trained exclusively on daytime annotations provided by the RGB daytime teacher model. We observe that the daytime-nighttime domain gap is the smallest for V1, while V2 and V3 suffer from a larger domain gap of the RGB modality, but achieve a higher daytime performance. 

Variants V4 and V5, are similar to variants V2 and V3, but with an additional domain discriminator, indicating that adding a domain discriminator loss to the overall training as described in Sec. \ref{sec:rgbtsemseg} greatly helps shrinking the domain gap within the RGB image modality. Variant V6, with active domain adaptation and two-stage training procedure as described in Sec.~\ref{sec:thermal_night_init} shows the best performance in both the daytime and the nighttime domain. We conclude that our proposed two-stage training scheme by first carrying out supervised training with two teachers and later fine-tuning with domain adaptation leads to the best results and helps aligning the feature representations between day and night as best as possible.


\section{CONCLUSION}
\label{sec:conclusion}
In this work, we presented a novel and robust approach for daytime and nighttime semantic segmentation of urban scenes by leveraging both RGB and thermal images. We showed that our HeatNet approach avoids expensive and cumbersome annotation of nighttime images by learning from a pre-trained RGB-only teacher model and by adapting to the nighttime domain. We further proposed a novel training initialization scheme by first pre-training our model with a daytime RGB-only teacher model and a nighttime thermal-only teacher model and subsequently fine-tuning the model with a domain confusion loss. We furthermore introduced a first-of-its-kind large-scale RGB-T semantic segmentation dataset, including a novel target-less thermal camera calibration method based on image gradient alignment maximization.
We presented comprehensive quantitative and qualitative evaluations on multiple datasets and demonstrated the benefit of the complementary thermal modality for semantic segmentation and for learning more robust RGB-only nighttime models.




{\small
\bibliographystyle{ieee_fullname}
\bibliography{egbib}
}

\end{document}